\documentclass{article} 
\usepackage{iclr2025_conference,times}

\iclrfinalcopy

\usepackage{amsmath,amsfonts,bm}

\def\eqref#1{equation~\ref{#1}}

\def\1{\bm{1}}

\DeclareMathAlphabet{\mathsfit}{\encodingdefault}{\sfdefault}{m}{sl}
\SetMathAlphabet{\mathsfit}{bold}{\encodingdefault}{\sfdefault}{bx}{n}

\usepackage{hyperref}
\usepackage{url}
\usepackage{graphicx}
\usepackage{subcaption}
\usepackage{amssymb}

\title{The Effect of Label Noise on the Information Content of Neural Representations}

\author{Ali Hussaini Umar\textsuperscript{1,*}, 
Franky Kevin Nando Tezoh\textsuperscript{1}, 
Jean Barbier\textsuperscript{2}\\
\textbf{Santiago Acevedo\textsuperscript{1}, 
Alessandro Laio\textsuperscript{1,2}}
\\
\textsuperscript{1}Scuola Internazionale Superiore di Studi Avanzati (SISSA), Trieste, Italy \\
\textsuperscript{2}International Centre for Theoretical Physics (ICTP), Trieste, Italy \\
\texttt{\{aumar,fnandote\}@sissa.it}, 
\texttt{jbarbier@ictp.it}, \texttt{\{sacevedo,laio\}@sissa.it}
}

\iclrfinalcopy 
\begin{document}

\maketitle

\begin{abstract}
In supervised classification tasks, models are trained to predict a label for each data point. In real-world datasets, these labels are often noisy due to annotation errors. While the impact of label noise on the performance of deep learning models has been widely studied, its effects on the networks' hidden representations remain poorly understood. We address this gap by systematically comparing hidden representations using the Information Imbalance, a computationally efficient proxy of conditional mutual information. Through this analysis, we observe that the information content of the hidden representations follows a double descent as a function of the number of network parameters, akin to the behavior of the test error. 
We further demonstrate that in the underparameterized regime, representations learned with noisy labels are more informative than those learned with clean labels, while in the overparameterized regime, these representations are equally informative. Our results indicate that the representations of overparameterized networks are robust to label noise. 
We also found that the information imbalance between the penultimate and pre-softmax layers decreases with cross-entropy loss in the overparameterized regime. This offers a new perspective on understanding generalization in classification tasks. 
Extending our analysis to representations learned from random labels, we show that these perform worse than random features.  This indicates that training on random labels drives networks much beyond lazy learning, as weights adapt to encode labels information.
\end{abstract}

\section{Introduction}
In real-world datasets used to train supervised machine learning models, label errors are unavoidably present, for example, because the data is annotated by humans. These errors in the ground-truth classification labels are commonly referred to as \emph{label noise} \cite{frenay2014comprehensive}, and pose a significant challenge to machine learning models. Several studies have reported its presence in numerous datasets, including those routinely used as benchmarks \cite{northcutt2021confident,northcutt2021pervasive,zhang2017method}. The presence of this noise has multiple consequences. Most notably, it degrades the generalization performance. Specifically, deep learning models tend to interpolate the training examples, regardless of whether the labels are correct or wrong. This unavoidably affects the performance on test datasets \cite{zhang2021understanding}. This effect has been theoretically studied for simple models such as linear and quadratic classifiers \cite{lachenbruch1966discriminant, mclachlan1972asymptotic, lachenbruch1979note} and empirically demonstrated in deep classifiers by \cite{zhang2021understanding, arpit2017closer}. Moreover, recent work by \citeauthor{nakkiran2021deep} reports that the emergence of the double descent phenomenon is strongly enhanced when label noise is present in the training data (see Sec. \ref{double_descent} for more discussion).

While the effect of label noise on the performance of neural networks is relatively well understood,  an important question remains open: How does label noise affect the \emph{hidden} representations of deep learning models? Do the representations trained with noisy labels differ significantly from those trained with clean labels? One might be tempted to guess that label noise should degrade the information content of a representation, but this notion, if correct, should be reconciled with a well-known fact: Hidden representation built with totally random weights but with sufficiently many neurons can interpolate virtually any function and be used to solve any classification problem \cite{rahimi2007random,goldt2020modeling,mei2022generalization,gerace2022probing}.
The properties of hidden representations are relevant because deep learning models are based on representation learning, and their success is generally attributed to their ability to learn relevant features from the input data \cite{lecun2015deep}. These features are encoded in the hidden representations. We believe that studying the effect of label noise on the quality of those features explicitly will contribute to a better understanding of the principles of feature learning and, by extension, deep learning.

In this study, we aim to characterize the hidden representations of neural networks trained on classification tasks under varying levels of label noise.
To investigate the quality of the features learned by the networks, we quantify the relative information content of different learned representations by estimating the Information Imbalance (II), a novel and computationally efficient statistical measure introduced in \cite{glielmo2022ranking}. II compares two different feature spaces for the same dataset (say space $A$ and space $B$), by analyzing how well the relative distances between samples are preserved when going from one space to the other. 
On a high level, it measures predictability from space $A$ to space $B$. For each point $i$, it identifies its nearest neighbor $j$ in space $A$. Then it calculates the mean rank of the distance between $i$ and $j$ in space $B$. A low mean rank indicates high predictability, as neighbors in $A$ remain neighbors in $B$. A high mean rank indicates randomness and a lack of predictability. We formalize this notion in Sec.~\ref{sec:II}. 
The II has solid theoretical grounds in terms of copula variables~\cite{glielmo2022ranking}, and remarkably, it was shown to be an upper bound of a conditional mutual information between random variables representing the pairwise distances in space $A$ and $B$ \cite{del2024robust}. 
%
Our key contributions are the following.
\begin{enumerate}
    \item [$\bullet$] We revisited the relationship between II and the conditional entropy between distances. We computed the explicit lower bound of II in the settings of one-dimensional embedding and a Gaussian denoising model, and show that the bound is tight when the embeddings are highly correlated and when the signal-to-noise ratio of the denoising model is high. 
    We demonstrate that it provides a manner to identify the "best" representation in a task-agnostic manner (Sec. \ref{II_RMI}). 
    \item [$\bullet$]By analyzing representation of neural networks trained on the MNIST and CIFAR-10 image classification tasks, we show that the double descent phenomenon observed in the test error is mirrored by a double descent in the information imbalance between hidden representations of identical networks trained with different initializations (Sec. \ref{Double Descent Phenomenon on the information content}). 
    \item [$\bullet$] We find that in the underparameterized regime, adding label noise to the training set yields more informative hidden representations compared to those learned with the ground-truth labels. In the overparameterized regime, hidden representations quickly become almost equally informative for moderate levels of label noise.
    Furthermore, label noise amplifies information loss between the hidden and the pre-softmax representations of a network. This suggests that label noise primarily degrades information in the last layer, while the quality of the feature encoding within hidden representations remains relatively preserved 
    (Sec. \ref{Label noise boosts information content in the underparameterized regime}). 
    \item [$\bullet$] In strongly overparameterized networks, representations learned with ground-truth labels and random features become mutually predictive. In contrast, representations learned with random labels contain less information than random features. The representation with lower information content gives worse test error 
    (Sec. \ref{Discriminating between random labels and random features}).
\end{enumerate}
Together, these results suggest that the information imbalance provides a proxy for comparing the quality of hidden representations in an unsupervised manner, that is, without utilizing the ground-truth labels to estimate its quality. 
\section{methodology}
\subsection{Information Imbalance }
\label{sec:II}
The Information Imbalance allows computing the relative information content of a representation space (say $A$) with respect to a second representation space (say $B$). The II is proportional to the average rank according to the distance in space $B$, denoted by $d^B$, conditioned on the nearest neighbors according to the distance in space $A$, denoted by $d^A$. In a sample of $N$ data points, the II is estimated by 
\begin{equation}
   \label{II}
    \Delta(A\rightarrow B) \sim \frac{2}{N} \langle r^B\;|\;r^A=0 \rangle =\frac{2}{N(N-2)} \sum_{i,j \;|\,r^A_{i,j}= 0} r_{i,j}^B,
\end{equation}
where $r_{i,j}^A \in \{0, 1,2,..., N-2\}$ ($r_{i,j}^B$ respectively) is the rank of the distance between data point $i$ and data point $j$ among all distances relative to the data point $i$. This ranking is obtained by ordering the distances from the smallest to the largest. For example, $r_{i,x}^A = 0$ if the data point $x$ is the nearest neighbor of the data point $i$ in space $A$.
The factor $\frac{2}{N}$ is introduced to ensure that if the two distances $d_A$ and $d_B$ are not correlated, then $\Delta(A\rightarrow B) \simeq 1$. 
If $\Delta(A\rightarrow B)\approx0$ and $\Delta(B \rightarrow A)\approx0$, then $A$ and $B$ are locally equivalent. On the other hand, if $\Delta(A\rightarrow B)\approx 1$ and $\Delta(B\rightarrow A)\approx 1$, then the information carried by $A$ and $B$ are independent or, equivalently, the two spaces are not predictive of each other. Because of the conditioning nature of the measure, the II is not symmetric in general; typically $\Delta(A\rightarrow B)$ is different from $\Delta(B\rightarrow A) $. If $\Delta(A\rightarrow B)<\Delta(B\rightarrow A) $, then space $A$ is more informative than space $B$ \cite{glielmo2022ranking}. The main difference between the II and other similarity measures is its asymmetry, which, as we will see, allows assessing whether one representation is more informative than another.  

It's worth mentioning that the II inherently satisfies the invariance properties expected for a robust similarity measure of neural network representation: Invariance to \textit{isotropic scaling} and \textit{orthogonal transformations} \cite{kornblith2019similarity}. It is straightforward to see that II satisfies these properties, as it depends only on the distribution of distances between features within each representation space. Since neither isotropic scaling nor orthogonal transformations alter the distribution of distances, it follows that $\Delta(\alpha A\rightarrow \beta B)= \Delta(A\rightarrow B)$ for any $\alpha,\beta \in \mathbb{R}$ and $\Delta(AU\rightarrow BV)= \Delta(A\rightarrow B)$ for any orthogonal matrices $U, V$. 
\subsection{Experimental Setup}\label{sec:setup}
We performed experiments on a two-baseline neural network for a classification task: A fully connected neural network (FCNN) and a standard convolutional neural network (CNN). The two networks have the following structure:
\begin{enumerate}
    \item 
A single-layer FCNN with $k$ hidden units, with input and output dimensions fixed to match the number of pixels in the input image and the number of classes in the task, respectively. 
\item  A family of  CNN consisting of $4$ convolutional stages of width $ [k, 2k, 4k, 8k]$, where $k$ is the width parameter, followed by a fully connected layer as a classifier. The Maxpoll is $[2,2,2,4]$.  For the entire convolution layer, the kernel size $=3$, stride $=1$, and padding $=1$. This architecture is identical to the one considered in Refs. \cite{gu2024unraveling} 
\end{enumerate}
The networks are trained using the mini-batch SDG algorithm to minimize the cross entropy loss. A batch size of 128 samples is used.  For FCNNs, we set the learning rate to start with $l_0=0.1$ and after every successive $50$ epoch it changes to $l_r = l_0/(\sqrt{1 + epoch/50})$. While for CNNs, the learning rate starts with $l_0=0.05$, and updates as $l_r = l_0/(\sqrt{1 + epoch\times50})$ after every $50$ epoch. This learning rate scheme was used in a recent paper by \cite{gu2024unraveling}.
The FCNNs are trained for image classification tasks with the MNIST dataset, and the CNNs are trained for image classification tasks with the CIFAR-10 dataset. For FCNNs, we follow the setup in Ref.~\cite{belkin2019reconciling}. This experiment involved training the neural network using $N_{train} = 4 \times 10^3$ samples from the MNIST training set and evaluating performance on the entire $N_{test} = 10^4$ MNIST test set.  For CNNs, we used all the training and test samples of CIFAR-10 ($N_{train} = 5\times 10^4$, $N_{test} = 10^4$) for training and evaluation. 
\section{Results}
We will first present analytical results related to the relationship between the II and a more standard information-theoretic quantity defined in terms of a Kullback–Leibler divergence. In particular, we will show that on a Gaussian models, the information-theoretic quantity can be computed analytically, and the former is an upper bound to the latter, with a gap between the two quantities that decreases as the two representation spaces tend to align. We will then present empirical results of our analysis on the hidden representations of FCNN and CNN. 

We conducted experiments with different types of labels. First, we used the ground-truth labels (equivalent to zero label noise, which is referred to as clean labels). Second, we introduced label noise by randomly changing the ground-truth label with probability $p$ to a random label, such that on average there are $N(1-p)$ ground-truth labels and $Np$ random labels.  We call such perturbed labels as noisy labels.  Note that the label noise procedure was applied once to the training set labels and remained fixed throughout training. The test set labels were always kept clean. 
In this setting, our objective is to quantify the relative information content between representations of different networks with identical architecture and representations from distinct layers of a trained network.
\subsection{Information Imbalance and its Lower bound for Gaussian Variables}
\label{II_RMI} 
\begin{figure}
    \centering
    \includegraphics[width=0.8\linewidth]{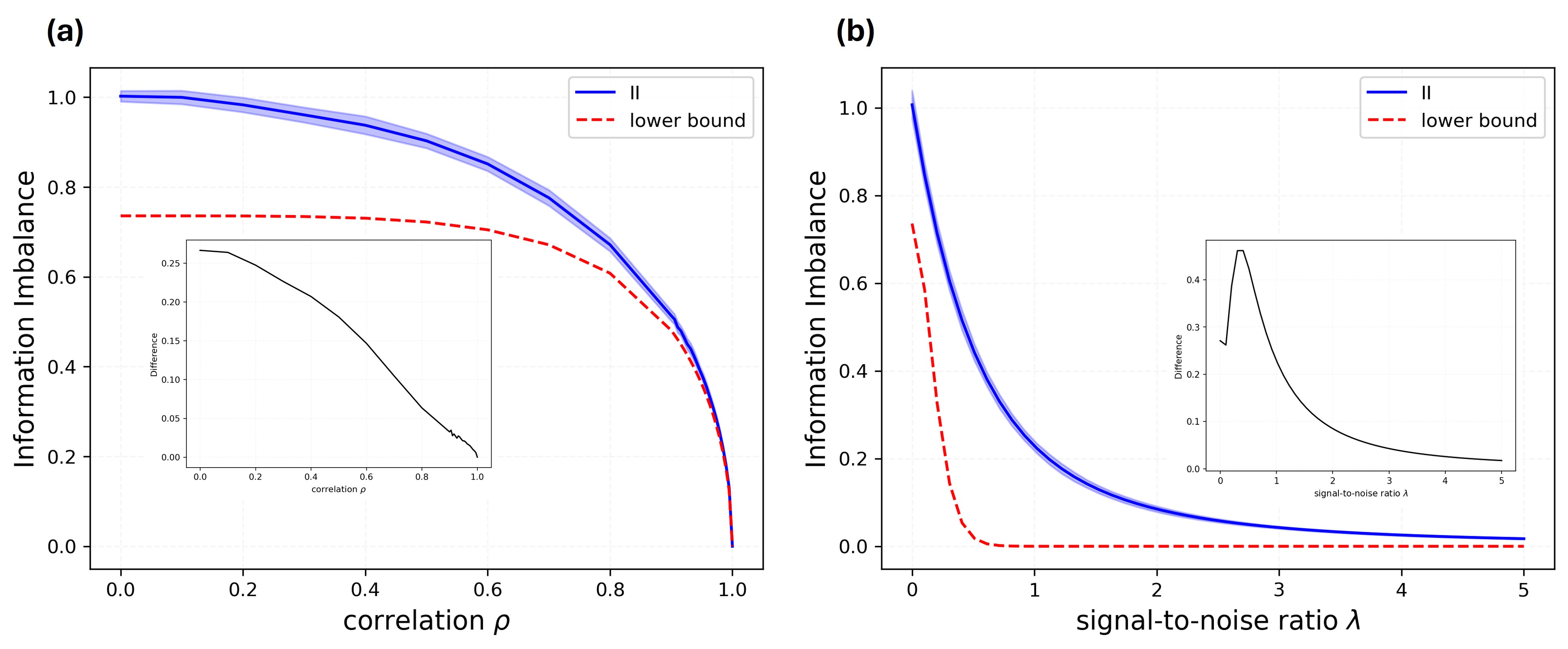}
    \caption{\textbf{left}: Information Imbalance (II) and its lower bound between the representation $x$ and $y$ representation of a Gaussian random variable $z$ as a function of correlation parameter $\rho$. \textbf{Right}: II and its lower bound for predicting the observed variable $\textbf{y}$ given the ground truth signal $\textbf{x}$ of a Gaussian denoising model as a function of signal-to-noise ratio $\lambda$, the dimension of the signal is $100$. The figure inside the main figure shows the difference between the II and its lower bound. The II curve is the average result of $30$ experiments computed on $10^3$ samples.}
    \label{fig:II&lower_bound}
\end{figure}
The inequality established by \cite{del2024robust} shows that II is an upper bound to a specific information-theoretic quantity called restricted mutual information (see Sec. \ref{RMI_definition} for its definition). The inequality is as follows:
\begin{equation}
    \label{inequality_main}
    \Delta(x \rightarrow y) \gtrsim 2\exp\left( -\lim_{\eta  \rightarrow 0} D_{\mathrm{KL}}[p(r_y| r_x = \eta ) \parallel p(r_y)]-1 \right),
\end{equation} 
where the random variable $r_x$ ($r_y$ respectively) denotes the pairwise distance in space $x$.
Computing the lower bound for generic data is very challenging, and thus in this work, we compute it for the first time in two simplified scenarios, allowing us to study the tightness explicitly.
First, we consider two one-dimensional representations of a random variable $\textbf{z}\sim \mathcal{N}(\textbf{0}, \mathbb{I}_d)$ that are defined as $x = \textbf{w}^T\textbf{z}$ and $y = \textbf{m}^T\textbf{z}$, where $\textbf{w},\textbf{m}\in \mathbb{R}^d$ are fixed vectors. In this setting, we computed the joint and marginal densities of the following random variables.
\begin{equation*}
    r_x = |x-x'| \quad\text{and}\quad   r_y = |y-y'|,
\end{equation*} where $x, x' \sim \mathcal{N}(0, \textbf{w}^T\textbf{w})$ and $y, y' \sim \mathcal{N}(0, \textbf{m}^T\textbf{m})$. We then compute the $D_\mathrm{KL}$ term in \eqref{inequality_main} using a first-order approximation (see Sec. \ref{dist_of_dist} for details of the computation):
 \begin{equation}
 \lim_{\eta  \rightarrow 0} D_{\mathrm{KL}}[p(r_y| r_x = \eta ) \parallel p(r_y)] = -\frac{1}{2}\log(1 - \rho^2)  - \frac{1}{2}\rho^2,
\end{equation}
where $\rho = \textbf{w}^T\textbf{m}/||\textbf{w}||||\textbf{m}||$ is the correlation between $x$ and $y$.
The expression depends only on the correlation parameters $\rho$.
Figure~\ref{fig:II&lower_bound}-(a) shows the plot of II and the second member in \eqref{inequality_main} as a function of $\rho$.  When $\rho = 0$ (independence), $D_{\mathrm{KL}}= 0$, the lower bound is $2e^{-1} \approx 0.74$ and $\Delta\rightarrow1$, indicating that $x$ and $y$ are not predictive of each other. When $|\rho|\rightarrow 1$, the II and its lower bound tend to zero, and the gap between them decreases linearly, indicating increasing alignment.
Second, we consider the Gaussian denoising model that is defined as \begin{equation}
\textbf{y} = \sqrt{\lambda}\textbf{x} +\textbf{e},
\end{equation}
where $\textbf{x}\in \mathbb{R}^d$ (ground truth signal) is a vector (or a scalar) drawn from $\mathcal{N}(0, \mathbb{I}_d)$, $\textbf{e}$ is addictive Gaussian noise, $\lambda\geq 0$ is the signal-to-noise ratio, and $\textbf{y}$ is the observed variable. 
Following the procedure above, the $D_\mathrm{KL}$ is 
\begin{equation}
 \lim_{\eta  \rightarrow 0} D_{\mathrm{KL}}[p(r_\textbf{y}| r_\textbf{x} = \eta ) \parallel p(r_\textbf{y})] = \frac{d}{2}\left(\log(1 + \lambda) - \frac{\lambda}{1 + \lambda}\right).
\end{equation}
Also, the result here depends only on the signal-to-noise ratio. Figure~\ref{fig:II&lower_bound}-(b) presents the results of II and its lower bound as a function of $\lambda$. In this setting, the II values for predicting the observed variable given the signal decrease monotonically as $\lambda$ increases, and the gap between the II and its lower bound decreases accordingly. This monotonic decrease in II is also compatible with the minimum mean square error behavior when estimating the signals from the observed variables, where the optimal estimation is achieved in an infinite value of $\lambda$ \cite{guo2005mutual}. This provides numerical evidence that the II is a proxy for an information-theoretic quantity that measures relative information content between distance spaces. Especially in the informative regime (when $\Delta(x \rightarrow y) \rightarrow 0$), II provides a better estimate of the right-hand side of \eqref{inequality_main}.
\subsection{Double Descent Phenomenon on the information content}
\label{Double Descent Phenomenon on the information content}
\begin{figure*}
    \centering
    \includegraphics[width=0.8\linewidth]{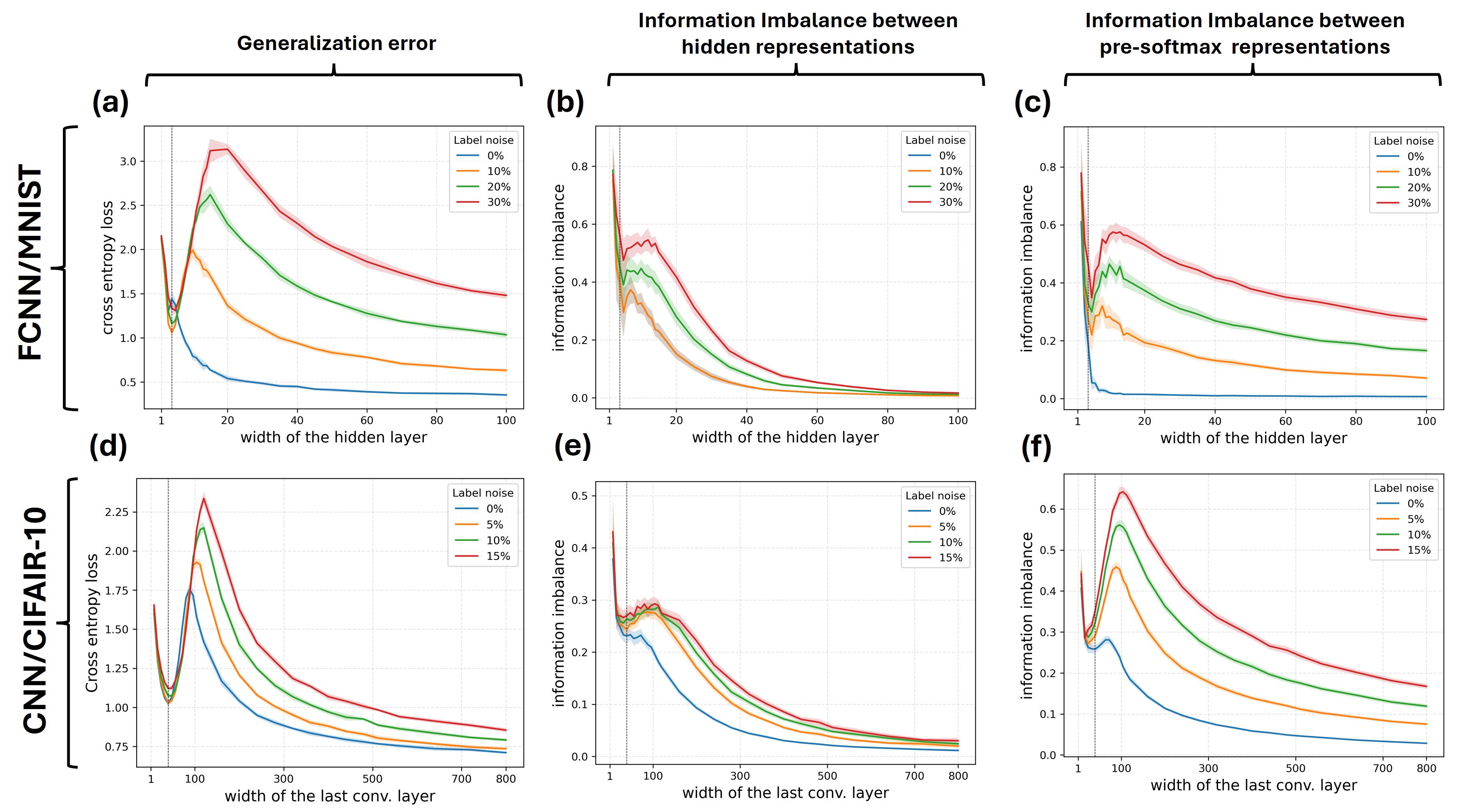}
    \caption{\textbf{Double descent phenomenon in the relative information content of statistically independent representations.} The top panels refer to results for single-layer FCNNs trained on the MNIST dataset, and the bottom panels correspond to results for CNNs trained on the CIFAR-10 dataset. All the curves in all the panels are plotted as a function of network parameters. Panels (a) and (d) show the generalization error for different levels of label noise. Panels (b) and (e) show the Information Imbalance between the hidden representations of identical networks trained on the same dataset with independent initializations. Panels (c) and (f) show the Information Imbalance between the pre-softmax representations (logits that provide the network output) of these networks.
    The Information Imbalance was computed using $2\times10^3$ test samples, with each curve representing an average over $90$ pairs of representations.}
    \label{fig:figure2}
\end{figure*}
The main goal of this work is to use a task-agnostic statistical measure, the Information Imbalance (II), to quantify the information content of representations learned by neural networks trained for classification tasks. 
Specifically, we consider two well-studied problems, the recognition of numbers in the MNIST dataset and of objects in the CIFAR-10 dataset. We use a single layer FCNN for the MNIST task and a standard CNN for the CIFAR-10 task. 
In both cases, we train the networks by varying the number of parameters, which was controlled by the number of hidden units ($k$) for FCNNs and the value of the width parameter ($k$) for CNNs (see Sec. \ref{sec:setup} for a detailed description of the models). For both architectures, we refer to $k$ as the \emph{width parameter}.
Panels (a) and (d) of Figure~\ref{fig:figure2} show the generalization error as a function of the width parameter for the MNIST and the CIFAR-10 datasets. These curves show the classical double descent behavior. 
For both datasets, the higher the noise in the labels, the higher the peak in the generalization error \cite{nakkiran2021deep}. In the MNIST case, if the label noise is set to zero, the peak is very small. This residual peak can be attributed to a small implicit noise on the original MNIST labels (\cite{northcutt2021confident,northcutt2021pervasive}. 
We also notice that, in the underparameterized regime, training with label noise \emph{reduces} the error in the MNIST case and does not significantly affect the error in the CIFAR-10 case.

We now analyze the information content of the hidden representations learned by these networks. In Figure~\ref{fig:figure2}-(b) for MNIST and Figure~\ref{fig:figure2}-(e) for CIFAR-10, we plot the II between representations of networks that share the same architecture and training data, but which were trained with different random initializations. The II is averaged between $10\,\times\,9$ pairwise combinations of the $10$ trained networks' representation, and is plotted as a function of the width parameter, as in the panels (a) and (d) of Figure~\ref{fig:figure2}. Remarkably, the II behavior closely resembles the double descent phenomenon observed in the generalization error. For the curves with label noise, the first minimum in the II  occurs at approximately the same width for which the generalization error is minimum. For the MNIST case, in the absence of label noise, the II peak is almost absent, which is consistent with what was observed for the generalization error. 
This analysis show that when networks begin to overfit the training data and the test error grows as a function of the width, their representations become less mutually predictive of each other. Mutual predictivity increases in the second branch of the double descent curves.

We observe one important difference between the test errors and the IIs: In the overparameterized regime, the II decays to zero for any label noise, while the error does not.  This indicates that overparameterized networks learn mutually predictive features despite being trained with different label noise levels. 
In the next Section, we will show that the hidden representations learned in the presence of moderate label noise are not only equivalent in statistically independent trainings, but also equivalent to those learned in the absence of noise. This suggests that the hidden representations, in the networks we considered, are (approximately) noise-independent in the strongly overparameterized regime.   

In Figure~\ref{fig:figure2} panels (c) and (f), we plot the II between the representations preceding the softmax, which provides the output, which in both cases consists of 10 learned features (the number of classes). Remarkably, in this case, the representations learned in statistically independent instances become approximately equivalent when the network is large \emph{only in the absence of label noise}. In the presence of noise, even if the networks are trained on the same (noisy) labels, the II between the pre-softmax activations increases to a level that is controlled by label noise. While the hidden representations are approximately noise-agnostic, the pre-softmax activations are strongly influenced by noise, even for strongly overparameterized networks. This suggests that the II between pre-softmax activations provides a proxy of the intrinsic level of noise in a dataset. Importantly, this proxy can be computed without having access to the uncorrupted labels and remarkably without computing the test error.  
\subsection{Label noise boosts information content in the underparameterized regime}
\label{Label noise boosts information content in the underparameterized regime}
\begin{figure*}
    \centering
    \includegraphics[width=0.8\linewidth]{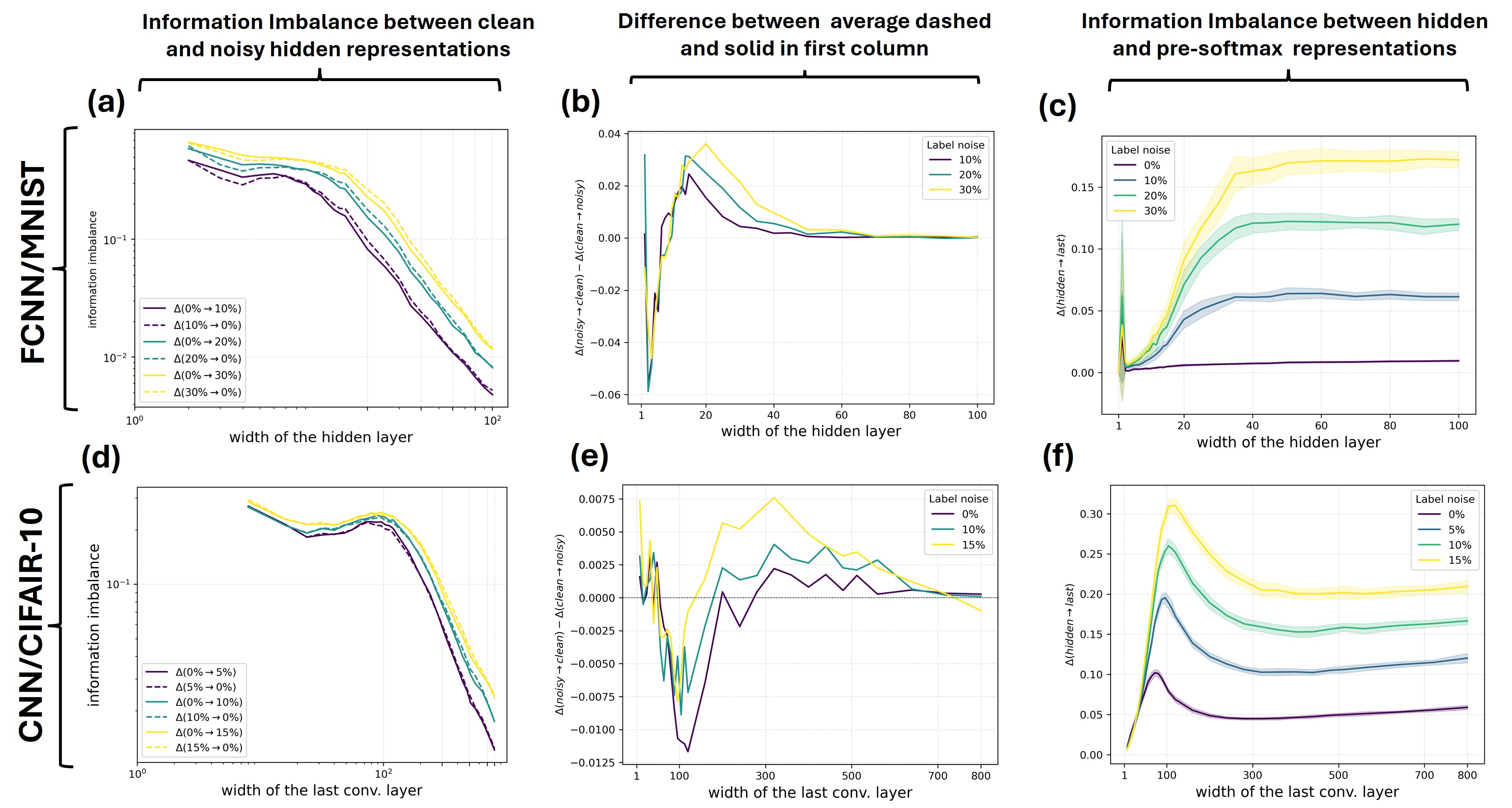}
     \caption{\textbf{Information loss between penultimate and last layer pre-softmax representations of overparameterized networks}: The top panels display results for single-layer FCNNs trained on the MNIST dataset, while the bottom panels show the results for CNNs trained on the CIFAR-10 dataset. All the curves in all the panels are plotted as a function of the number of network parameters. Panels (a) and (d) present the Information Imbalance (II) curves between hidden representations of identical networks trained with and without label noise. Panels (b) and (e) show the difference between the two IIs in panels (a) and (d), respectively. i.e, $\Delta(\cdot\% \rightarrow 0\%) - \Delta(0\% \rightarrow \cdot\%)$ . Panels (c) and (f) show the II for predicting the last layer preactivations representation given the hidden (penultimate) representation of a network trained with a certain label noise ratio. For FCNN (respectively CNN), the curves are averaged over $20$ (respectively, $10$) independent networks trained with different initializations.}
    \label{fig:figure3}
\end{figure*}
In the previous Section, we compared the representations of networks with identical architectures trained on the same dataset, with the only difference being the initialization of their weights. Here, we examine the relative information content of the representations learned with different label noise. In particular, we compare the representations of networks trained with clean and noisy labels. Panels (a) and (d) of Figure~\ref{fig:figure3} show the II between the representations learned with clean and noisy labels for FCNN and CNN, as a function of the width parameter.  The II decays to zero approximately with a power law both in the FCNN trained on the MNIST dataset (Figure~\ref{fig:figure3}-(a)) and in the CNN trained on the CIFAR-10 dataset (Figure~\ref{fig:figure3}-(a)). This implies that in the highly overparameterized regime, the representations of networks trained with and without label noise are asymptotically predictive of each other. As anticipated in the previous Section, the hidden representations of the networks we considered are only marginally affected by the noise in the strong overparameterized regime.

Importantly, the II between the representations learned with clean labels and with noisy labels is asymmetric. The continuous line corresponds to $\Delta(\text{clean} \rightarrow \text{noisy})$, while the dashed line corresponds to $\Delta(\text{noisy}\rightarrow \text{clean})$. In panel (b) for the MNIST case and panel (d) for the CIFAR-10 case in Figure~\ref{fig:figure3}, we plot the \emph{difference} between II in the two directions, $( \Delta(\text{noisy} \rightarrow \text{clean}) - \Delta(\text{clean} \rightarrow \text{noisy}))$. This difference serves as a measure of information asymmetry. A negative difference indicates that the noisy representation is more informative than the clean representation. This happens in both the MNIST (panel (b)) and the CIFAR-10 (panel (e)) cases when the width parameters are small, and the networks are in the "classical" regime. In this regime, adding a small label noise helps in developing representations that are marginally more informative. This is consistent with the observation that the test error in this regime is smaller in the presence of label noise for MNIST  (Fig.~\ref{fig:figure2}-a) and independent of the label noise for CIFAR-10 (Fig.~\ref{fig:figure2}-b). Indeed, it is a well known finding that in this classical regime, the loss landscape is glassy \cite{geiger2019jamming, baity2018comparing} and that performing training in the presence of noise helps regularize it \cite{keskar2016large}.

When the number of network parameters exceeds the number of training examples, the network enters the "modern" regime, corresponding to the second branch of the double descent curve. In this regime, the II swaps sign, implying that the clean representation becomes more informative than the noisy representation. As the number of parameters is large, the landscape loses its glassy characteristics \cite{baity2018comparing}. Instead, the network gains enough parameters to interpolate any training labels, including the wrong ones.  In these conditions, adding a label noise reduces the information content of the representations. When the number of parameters is very large, the noisy and clean representations become approximately equally predictive within our statistical accuracy. In our interpretation, the representations become similar to random features, which are known to be adequate for solving any interpolation problem \cite{mei2022generalization}. We will discuss this connection in detail in the next Section. 

This analysis, along with results presented in the previous Section, suggests that when a network is strongly overparameterized, its hidden representations become largely insensitive to noise and equally informative to the representations learned with clean labels. This naturally raises the question: Where does the gap in the test error originate? To address this question, we examine how the hidden and the pre-softmax representations of a network predict each other. In Panels (c) and (f) of Figure~\ref{fig:figure3}, we plot the II of the hidden representation predicting the pre-softmax representation for different levels of label noise, averaged over the random initialization of the networks. For clean labels, the II is close to zero, especially in the MNIST case, indicating that there is high information sharing between these representations. Instead, if the labels are partially corrupted, the II becomes significantly different from zero even in strongly overparameterized networks. This indicates that, the task leads to information loss, induced by the fact that the task is (partially) misspecified. It is pretty remarkable that in CNNs, which have four hidden layers, this information degradation happens only for the representation immediately before the output, while the penultimate, as shown in panels (d) and (e) of Figure~\ref{fig:figure3}, remains largely noise-insensitive.
\begin{figure*}
    \centering
    \includegraphics[width=0.75\linewidth]{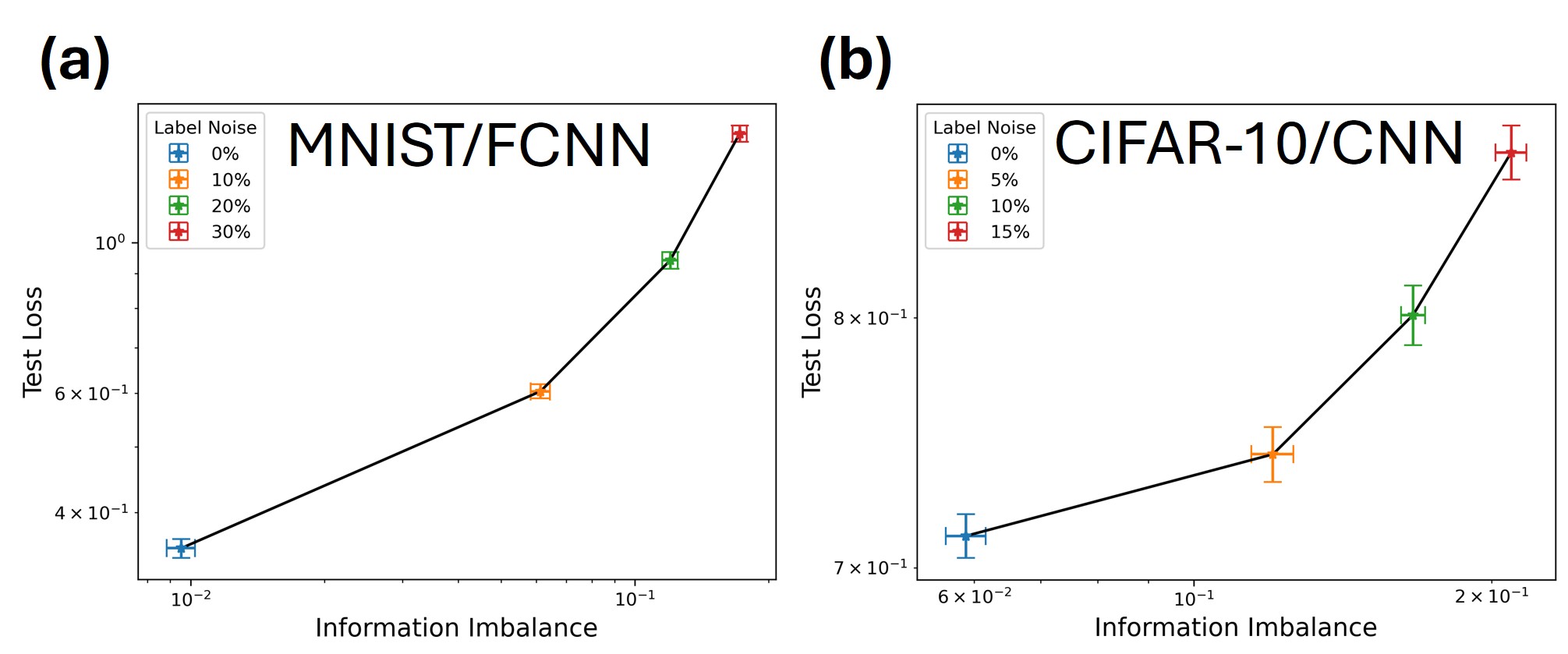}
    \caption{\textbf{Performance of trained network against the information imbalance of its hidden representation predicting its pre-softmax representation}. Left: Results of a single-layer FCNN with 100 hidden units trained on the MNIST classification task. Right: Results for a vanilla CNN with a 100-width parameter trained on the CIFAR-10 classification task. Each point is the average result of $20$ networks trained with independent initialization.}
    \label{fig:II_CE}
\end{figure*}
On the other hand, when we compare the II between hidden and pre-softmax representations of overparameterized networks and their performance, we found a clear trend: Good network performance correlates with lower II values. Specifically, the lower the cross entropy loss, the lower the II (see Figure~\ref{fig:II_CE}). This observation suggests that having a lower II between these representations is a sufficient condition for a network to generalize well. Based on this, we argue that II can be used as an unsupervised tool to understand generalization in a classification task. 
\subsection{Does training with random labels lead to random features?}
\label{Discriminating between random labels and random features}
\begin{figure}
    \centering
    \includegraphics[width=0.8\linewidth]{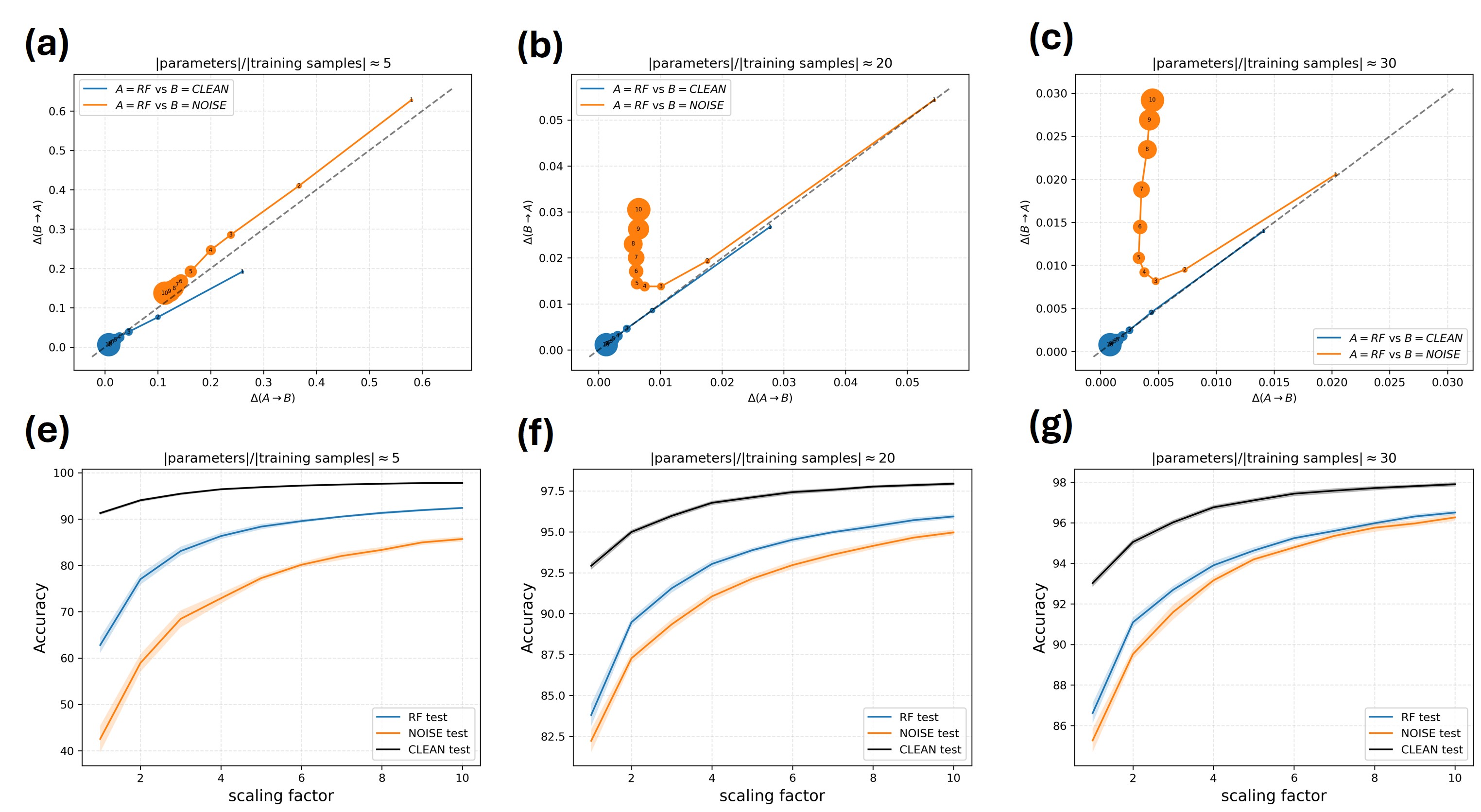}
    \caption{\textbf{Effect of label Memorization in representation}: The first row shows the evolution of information imbalances between two representations as network parameters $(P)$ and the training sample size $(N)$ are scaled proportionally by a scaling factor $c$, while their ratio $\alpha =P/N$ remains fixed. Figures (a), (b), and (c) correspond to cases where $\alpha \approx 5$,$\alpha \approx 20$, and $\alpha \approx 30$, respectively. The blue curves represent results between random features (RF) and representation learned with clean labels (CLEAN), while the orange curves represent results between RF and representation learned with random labels (NOISE). Initially, $N_0=4000$, and $P_0=\alpha N_0$  (smallest circle, $c=1$), and both $N$ and $P$ are scaled up proportionally to the largest circle ($c=10$), where $N=10N_0$ and $P=\alpha 10N_0$.  The second row displays the test accuracy of a classifier trained on a clean task using the RF, CLEAN, and NOISE representations.  Panels (e), (f), and (h) correspond to the representations in Panels (a), (b), and (c), respectively. All the curves are average results of 20 independent experiments on the MNIST data set.}
    \label{fig:rtest}
\end{figure}
In the previous sections, we have seen that hidden representations of overparameterized networks tend to become equally informative, despite the presence of noise in the training set. This is not a surprise: It is well known that the so-called random feature model, namely a neural network in which the weights are fixed at random, can solve with arbitrary accuracy any regression or classification problem, if the hidden layer is sufficiently wide \cite{mei2022generalization, guth2024rainbow}. Moreover, it is known that a deep learning model can interpolate any set of random labels \cite{zhang2021understanding}, which, in the language of this work, is equivalent to setting $p=1$. However, one can wonder if the representations learned by training the model with strong label noise are equivalent to random features.  
To explore this, we analyzed three representations: (RF) the representation of a random feature model, (CLEAN) a representation learned by training the network on clean labels, and (NOISE) a representation learned by training the network on random labels, namely with $p=1$. We studied how CLEAN and NOISE representations predict the RF as we proportionally increase the number of network parameters $P$ and the number of training data $N$ by a factor $c$, while keeping their ratio $\alpha =  P/N$ fixed.

Figure~\ref{fig:rtest} presents the plots of the two IIs ($\Delta(A\rightarrow B) $ and $\Delta(B\rightarrow A) $) against each other, for different values of $\alpha$ ranging from weak overparameterization ($\alpha=5$, left) to strong overparameterization ($\alpha=30$, right). The blue curves show the evolution of the IIs between the RF and the CLEAN as a function of the factor $c$. Initially, $N_0=4000$, and $P_0=\alpha N_0$  (smallest circle), then we scale both $N$ and $P$ with $c$ until the largest circle, where $N=10N_0$ and $P=\alpha 10N_0$. 
For all three levels of overparameterization shown and for any value of $N$ (and $P$), the points are approximately on the diagonal, namely $\Delta(CLEAN\rightarrow RF) \simeq \Delta(RF\rightarrow CLEAN)$. This implies that the two representations are equally informative \cite{glielmo2022ranking}. The larger the $c$, the more the IIs approach zero, the value corresponding to equivalent representations \cite{glielmo2022ranking}.
The orange curves show the evolution of the IIs between the NOISE and the RF representation. At moderate overparameterization (panel a), the points lie approximately along the diagonal. For larger overparameterization (panels b and c), the curves follow the diagonal for small values of $c$, but deviate significantly for larger $c$. This indicates that under these conditions $\Delta(\text{NOISE} \rightarrow \text{RF}) > \Delta(\text{RF} \rightarrow \text{NOISE})$, and the representation learned with random labels becomes less informative than the representation of random features \cite{glielmo2022ranking}.

This analysis allows us to conclude that the representations learned by training the model with strong label noise are \emph{not} equivalent to random features. Moreover, the latter are of a higher quality, since they are asymptotically equivalent to the representations learned without noise in the context of II (blue curves). We perform an additional experiment to check the performance of a classifier trained on the ground-truth task using the RF, CLEAN, and NOISE representations. We consistently found that a classifier trained using the RF gives better test accuracy than a classifier using the NOISE representation. The results are shown in the second row of Figure~\ref{fig:rtest}.  Panels (e), (f), and (g) correspond to the cases where $\alpha \approx 5$,$\alpha \approx 20$, and $\alpha \approx 30$, respectively. 
However, we observed that the performance gain in using the RF over the NOISE representation decreases for larger values of $\alpha$ (see panels (e) and (f)). In these strongly overparameterized conditions, the II is very small (meaning that the representations are almost equivalent). As we discussed earlier, this is a hallmark of convergence and can be used to predict that the test error will be extremely small. This is indeed what we observed in Panel (f). We notice that even under these conditions, a performance gap between the NOISE and RF models remains, consistent with the asymmetry of the II. This suggests that networks trained with random labels encode information about those labels within their representations, inducing a bias that persists even when $N$ and $P$ are large.

\section{Conclusion and Discussion}
In this work, we examined how label noise affects the information content of the learned representations of neural networks. 
We measure the information content using Information Imbalance (II), which was shown theoretically to be an upper bound on a conditional mutual information~\cite{del2024robust}. We compute explicitly the lower bound of II for two Gaussian models and prove empirically that II provides an excellent estimate of its lower bound when the representations are closely similar. 
Varying model sizes, we show that the double descent phenomenon emerges in the information content of hidden (penultimate) representations of identical networks trained on the same data with different initializations. This highlights that as networks begin overfitting and test error increases with width, their representations become less mutually predictive.
We also show that, in the overparameterized regime, the representations learned without and with a small level of label noise are equally informative of each other, indicating that these representations are largely unaffected by the label noise. However, we observe information loss between the penultimate and the pre-softmax final representations, which increases with the percentage of label noise in the training set. This information loss also predicts the network performance: networks with small cross entropy loss exhibit small information loss. 
Based on this finding, we believe that II can be used as a scalable unsupervised tool to study generalization in neural networks. 
\subsection{Limitations}
Throughout this paper, we analyzed the representations of two baseline networks for classification tasks: a single-layer fully connected neural network (FCNN) and a vanilla convolutional network (CNN). The FCNNs are trained on the MNIST dataset, while the CNNs are trained on the CIFAR-10. Clearly, it would be very interesting to extend the analysis to more realistic data sets (say, Imagenet) and architectures (say, a state-of-the-art transformer for image classification). We focused on baseline models to lay the groundwork for understanding how label noise affects hidden representations and to interpret the representations underlying double descent. Despite the simplicity of this setting, we obtained nontrivial insights into the internal mechanisms of those models when trained with label noise. 
\subsection{Future Direction}
On the theoretical side, we would like to better understand the behavior of II and its lower bound in high-dimensional spaces, including cases where the dimensions of the two spaces that are compared in the II differ. Extending the analysis in this direction would allow interpreting more rigorously the results observed in hidden representations, which are very high-dimensional. 
Another promising direction is to study the impact of noise in natural language processing tasks, to examine whether the patterns observed in vision networks' representation also hold in language models.  

\subsubsection*{Acknowledgments}
We thank Sebastian Goldt and Vittorio Del Tatto for insightful discussions and suggestions.
A.L. and S.A. acknowledge financial support by the region Friuli Venezia Giulia (project F53C22001770002). 
J.B. was funded by the European Union (ERC, CHORAL, project number 101039794). Views and opinions expressed are however those of the authors only and do not necessarily reflect those of the European Union or the European Research Council. Neither the European Union nor the granting authority can be held responsible for them.

\bibliography{iclr2025_conference}
\bibliographystyle{iclr2025_conference}
\newpage
\appendix
\section{Appendix}
\subsection{About Double Descent}
\label{double_descent}
Classical statistical theory suggests avoiding models with too many parameters to avoid overfitting and poor generalization to new test examples. According to this theory, it is preferable to keep the number of parameters (model complexity) of the model below the interpolation threshold (the point at which the model perfectly fits the training data). This helps to build a model with a small generalization error. However, modern neural networks often have far more parameters than training examples, allowing them to fit the training data perfectly. These models are so complex that they can fit the training data even if the actual labels were replaced with random ones \cite{zhang2021understanding}. Despite this, these models often achieve low generalization errors and outperform models trained according to the classical theory.

The double descent curve justifies this behavior, explaining why deep neural networks perform well. Initially, the curve follows the classical U-shaped risk pattern up to the interpolation threshold. Beyond this point, the test error starts to decrease again, and frequently a global minimum is achieved in the second part of the curve. Specifically, the minimum error often appears in infinite complexity. The second part of the double descent curve is referred to as the 'modern' interpolation regime \cite{belkin2019reconciling}.

The original work where the double descent phenomenon was presented is \cite{belkin2019reconciling} and was revealed on the scale of model size. However, a follow-up work by \cite{nakkiran2021deep} has also shown a similar behavior of the generalization curve as a function of the size of the training set and training time. In addition, the research reports that all forms of double descent strongly emerge in settings with label noise in the training set. Moreover, they find that increasing label noise significantly amplifies the generalization error peak around the interpolation threshold. i.e, the peak amplitude correlates positively with the label noise ratio in the training data.
\subsection{Information Imbalance as upper bound of information Theoretic quantity}
\label{RMI_definition}
\subsubsection{Information Content Between Representation Space}
The relative information content of the two distance spaces is measured by mutual information $\text{I}(r_x, r_y)$. However, mutual information will not be useful in quantifying how well a certain representation space predicts a geometric property of another representation space due to its symmetric nature. \cite{del2024robust} proposed an asymmetric information-theoretic quantity called restricted mutual information, which they define as 
 \begin{equation*}
\text{I}^\epsilon(r_x \rightarrow r_y) = \int_0^\eta\int_0^\infty dr_xdr_y p(r_x, r_y) \log\left(\frac{p(r_x, r_y)}{p(r_x)p(r_y)}\right) 
\end{equation*} 
where $\eta = F^{-1}_x(\epsilon)$ and $F_x(r) = \int_0^r p(r_x) dr_x$ denote the cumulative distribution function of $r_x$. This quantity is proposed based on the intuition that a good distance space should be locally semantically meaningful. i.e, a space where a data point is enclosed by points that are "similar". Therefore, in the limit of small $\epsilon$, the restricted mutual information quantifies the shared information by the random variables $r_x$ and $r_y$ when conditioned on the events $d = \eta $. With the assumption that large distances are not informative, the authors (\citeauthor{del2024robust}) interpret this measure as the actual information in $r_y$ that is contained in $r_x$. By expanding the restricted mutual information in the limit $\epsilon \rightarrow 0$, the first nonzero term is its first derivative around zero:
\begin{align*}
  \lim_{\epsilon \rightarrow 0} \frac{\text{I}^\epsilon(r_x \rightarrow r_y)}{\epsilon}&=\frac{\partial}{\partial \epsilon}\text{I}^\epsilon(r_x; r_y)|_\epsilon\\&=\frac{1}{p(r_x = \eta )}\frac{\partial}{\partial r_x}\text{I}^\epsilon(r_x; r_y)|_\eta \\ &= \frac{1}{p(r_x = \eta  )}\frac{\partial}{\partial r_x}\bigg[\int_0^c\int_0^\infty dr_x dr_y p(r_x, r_y) \log\left(\frac{p(r_x, r_y)}{p(r_x)p(r_y)}\right)\bigg] \bigg|_\eta 
  \\&= \frac{1}{p(r_x = \eta )}\frac{\partial}{\partial r_x}\int_0^c dr_x p(r_x) \bigg[ \int_0^\infty dr_y p(r_y| r_x) \log\left(\frac{p(r_y|r_x)}{p(r_y)}\right)  \bigg]\bigg|_\eta \\
&=  \lim_{\eta  \rightarrow 0}\int_0^\infty dr_y p(r_y| r_x=\eta ) \log\left(\frac{p(r_y|r_x=\eta )}{p(r_y)}\right)
\end{align*}
Therefore, in the limit of $\epsilon \rightarrow 0$, the restricted mutual information is given by
\begin{equation}
     \lim_{\epsilon \rightarrow 0} \frac{\text{I}^\epsilon(r_x \rightarrow r_y)}{\epsilon} = \lim_{\eta  \rightarrow 0} D_{\mathrm{KL}}[p(r_y| r_x = \eta ) \parallel p(r_y)],
\end{equation}
where $D_{\mathrm{KL}}$ denotes the Kullback-Leibler divergence (a pseudo-distance). This divergence represents the amount of information (in nats) gained about $r_y$ when conditioning on $r_x = c$, compared to the marginal distribution of $r_y$. It quantifies how much the distribution of $r_y$ changes when conditioned on small distances in space $x$—the greater the change, the more informative space $x$ is with respect to space $y$.
\subsubsection{Relationship between Information Imbalance and Restricted Mutual Information}
The bridge between the information imbalance measure and restricted mutual information rests on the statistical theory of copulas. Given a two-dimensional random variable $r = (r_x, r_y)$ with joint distribution $p(r_x, r_y)$ and marginal densities $p(r_x)$ and $p(r_y)$. Its copula variable is define as
\begin{equation*}
    c = (c_x, c_y) = (F_x(r_x), F_y(r_y)), 
\end{equation*}
where $F_x(r_x)$ and $F_y(r_y)$ are the cumulative densities of random variables $r_x$ and $r_y $. According to Sklar's theorem \cite{nelsen2006introduction}, the distribution of $r = (r_x, r_y)$ can be decomposed as the product of the marginals and the joint copula distribution:
\begin{equation*}
    P(r_x, r_y) = p_c(c_x, c_y) p(r_x)p(r_y),
\end{equation*}
where $p_c(c_x, c_y)$ is the joint copula distribution. This relationship between the distribution of $r$ and its respective copula distribution allows us to rewrite the restricted mutual information as
\begin{align}\label{rmi}
   \lim_{\epsilon \rightarrow 0} \frac{\text{I}^\epsilon(r_x \rightarrow r_y)}{\epsilon}&= \lim_{\eta \rightarrow 0} \int_0^\infty dr_y p(r_y\;|\; r_x =\eta) \log\left(\frac{p(r_y \;|\; r_x=\eta)}{p(r_y)}\right)\\
   &= \lim_{\epsilon \rightarrow 0} \int_0^1 d c_y p_c(c_y\;|\; c_x =\epsilon) \log p_c(c_y \;|\; c_x=\epsilon)\\
   &= -\lim_{\epsilon \rightarrow 0} H(c_y | c_x = \epsilon).
\end{align}
In the work of \cite{del2024robust}, they show that the information imbalance is an upper bound of an information-theoretic measure:
\begin{equation}
\label{inequality1}
    \Delta(x \rightarrow y) \gtrsim 2 \exp\left(  \lim_{\epsilon \rightarrow 0} H(c_y | c_x = \epsilon)-1 \right) ,
\end{equation}
equivalently
\begin{equation}
\label{inequality2}
    \Delta(x \rightarrow y) \gtrsim 2\exp\left( -\lim_{\eta  \rightarrow 0} D_{\mathrm{KL}}[p(r_y| r_x = \eta ) \parallel p(r_y)]-1 \right).
\end{equation}
Details about the proof can be found in the supplementary material of their work. Due to the asymptotic relationship in establishing the bound,  the authors state that the information imbalance provides a better estimate of the right-hand side of \eqref{inequality2} in the informative regime, namely when $\Delta(x \rightarrow y) = 0$.  
\subsection{Distributions of Distance}
\label{dist_of_dist}
\subsubsection{1-dimensional Embedding}
Consider a $d$-dimensional random variable $\textbf{z}$ that follows $\mathcal{N}(\textbf{0}, \mathbb{I}_d)$ and let $x = \textbf{w}^T\textbf{z}$ and $y = \textbf{m}^T\textbf{z}$ be two distinct $1$-dimensional representation of $\textbf{z}$, where $\textbf{w},\textbf{m}\in \mathbb{R}^d$ are fixed vectors. The random variables $x$ and $y$ are themselves normally distributed. We are interested in the joint and marginal distributions of random variables $r_x$ and $r_y$ denoting the pairwise distances in $x$ and $y$, which are defined as 
\begin{equation*}
    r_x = |x-x'| \quad\text{and}\quad   r_y = |y-y'|,
\end{equation*} where $x, x' \sim \mathcal{N}(0, \textbf{w}^T\textbf{w})$ and $y, y' \sim \mathcal{N}(0, \textbf{m}^T\textbf{m})$.
The distribution of $r_x$ can be obtained from the following integral ($r_y$, respectively).
\begin{equation*}
    p(r_x) = \mathbb{E}_{x,x'}[\delta(|x-x'|- r_x)] = \mathbb{E}_{x}[p(r_x|x)],
\end{equation*}
where 
\begin{equation*}
    p(r_x|x)=  \mathbb{E}_{x'}[\delta(|x-x'|- r_x)] = \int_{-\infty}^{\infty}dx'p(x')\delta(|x-x'|- r_x)
\end{equation*}
Let assume $r_x >0$. Using the identity for the delta function that involves an absolute value:
\begin{equation*}
    \delta(|x-x'|- r_x) =  \delta((x-x')- r_x) + \delta((x-x') + r_x),
\end{equation*}
The expectation becomes:
\begin{align*}
    p(r_x|x)&=\int_{-\infty}^{\infty}dx'p(x')\delta((x-x') - r_x) + \int_{-\infty}^{\infty}dx'p(x')\delta((x-x') + r_x)\\
    &=\int_{-\infty}^{\infty}dx'p(x')\delta(-x'-(r_x -x)) + \int_{-\infty}^{\infty}dx'p(x')\delta(-x' - (-r_x - x))\\
    &= \frac{1}{\sqrt{2 \pi\sigma_x^2}}\exp{\bigg(-\frac{(r_x-x)^2}{2\sigma_x^2}\bigg)} + \frac{1}{\sqrt{2 \pi\sigma_x^2}}\exp{\bigg(-\frac{(r_x+x)^2}{2\sigma_x^2}\bigg)}
\end{align*}
Now, taking the expectation over $x$ gives the following.
\begin{align*}
    p(r_x)
    &= \int_{-\infty}^\infty dxp(x)\bigg[\frac{1}{\sqrt{2 \pi\sigma_x^2}}\exp{\bigg(-\frac{(r_x-x)^2}{2\sigma_x^2}\bigg)} + \frac{1}{\sqrt{2 \pi\sigma_x^2}}\exp{\bigg(-\frac{(r_x+x)^2}{2\sigma_x^2}\bigg)}\bigg]\\
    &= \frac{1}{\big(\sqrt{2 \pi\sigma_x^2}\big)^2}\bigg[\int_{-\infty}^\infty dx\exp{\bigg(-\frac{(r_x-x)^2+ x^2}{2\sigma_x^2}\bigg)} + \int_{-\infty}^\infty dx\exp{\bigg(-\frac{(r_x+x)^2 + x^2}{2\sigma_x^2}\bigg)}\bigg]\\
    &= \frac{1}{\big(\sqrt{2 \pi\sigma_x^2}\big)^2}\bigg[\exp{\bigg(-\frac{r_x^2}{4\sigma_x^2}\bigg)}\sqrt{\pi\sigma_x^2} +\exp{\bigg(-\frac{r_x^2}{4\sigma_x^2}\bigg)}\sqrt{\pi\sigma_x^2}\bigg]\\
    &= \frac{1}{\sqrt{\pi\sigma_x^2}}\exp{\bigg(-\frac{r_x^2}{4\sigma_x^2}\bigg)},
\end{align*}
Therefore,
\begin{equation}
\label{mar_x}
p(r_x)= \frac{1}{\sqrt{\pi\sigma_x^2}}\exp{\bigg(-\frac{r_x^2}{4\sigma_x^2}\bigg)}, 
\end{equation}
and 
\begin{equation}
\label{mar_y}
p(r_y)= \frac{1}{\sqrt{\pi\sigma_y^2}}\exp{\bigg(-\frac{r_y^2}{4\sigma_y^2}\bigg)}.
\end{equation}
The joint distribution for the random variables $r_x$ and $r_y$ is obtained by computing the following expectation:
\begin{equation*}
    p(r_x, r_y) = \mathbb{E}_{(x,y), (x',y')}[\delta(|x-x'|- r_x)\delta(|y-y'|- r_y)].
\end{equation*}
Let us define 
\begin{equation*}
    u= x-x'=w^T(z-z')\quad \text{and} \quad v= y-y'=m^T(z-z').
\end{equation*}
The vector $z-z'$ follows $\mathcal{N}(0, 2\mathbb{I})$, so $(u, v)$ is bivariate normal with mean $\textbf{0}$ and covariance matrix:
\begin{equation*}
     \Sigma 
= 2\begin{bmatrix}
\sigma_x^2 & \rho \sigma_x \sigma_y \\
\rho \sigma_x \sigma_y & \sigma_y^2
\end{bmatrix},
\end{equation*}
where $\sigma_x^2=\textbf{w}^T\textbf{w}$, $\sigma_y^2=\textbf{m}^T\textbf{m} $ and $\rho = \textbf{w}^T\textbf{m}/\sigma_x \sigma_y$. The joint probability distribution function of $(u, v)$ is:
\begin{equation*}
    f(u, r) = \frac{1}{4\pi \sigma_x \sigma_y \sqrt{(1-\rho^2)}}\exp\left(- \frac{u^2}{4\sigma_x^2(1-\rho^2)} + \frac{\rho uv}{2\sigma_x\sigma_y(1-\rho^2)}- \frac{v^2}{4\sigma_y^2(1-\rho^2)}\right).
\end{equation*}
The quantity we care about is:
\begin{equation*}
    \mathbb{E}_{(x,y), (x',y')}[\delta(|x-x'|- r_x)\delta(|y-y'|- r_y)] =  \mathbb{E}_{(u,v)}[\delta(|u|- r_x)\delta(|v|- r_y)]
\end{equation*}
For $r_x, r_y>0$, we expand the delta function:
\begin{equation*}
    \delta(|u|- r_x)\delta(|v|- r_y)= (\delta(u- r_x) +\delta(u  + r_x))(\delta(v- r_y) + \delta(v + r_y))
\end{equation*}
\begin{align*}
 \mathbb{E}_{(u,v)}[\delta(|u|- r_x)\delta(|v|- r_y)] =& \mathbb{E}_{(u,v)}[\delta(u - r_x)\delta(v- r_y)] + \mathbb{E}_{(u,v)}[\delta(u- r_x)\delta(v + r_y)] + \\&\mathbb{E}_{(u,v)}[\delta(u + r_x)\delta(v - r_y)] + \mathbb{E}_{(u,v)}[\delta(u + r_x)\delta(v+ r_y)]\\
 =& f(r_x, r_y) + f(-r_x, r_y) + f(r_x, -r_y) + f(-r_x, -r_y)\\
  =& 2f(r_x, r_y) + 2f(r_x, -r_y)
\end{align*}
Therefore,
\begin{equation}
\label{joint_xy}
 p(r_x, r_y)= \frac{1}{
 \pi \sigma_x \sigma_y \sqrt{(1-\rho^2)}}\exp\left(- \frac{r_x^2}{4\sigma_x^2(1-\rho^2)} - \frac{r_y^2}{4\sigma_y^2(1-\rho^2)}\right)\cosh\left(\frac{\rho r_x r_y}{2\sigma_x\sigma_y(1-\rho^2)}\right).
\end{equation}
\subsubsection{Gaussian Denoising Model}
Consider the general denoising model
\begin{equation}
\textbf{y} = \sqrt{\lambda}\textbf{x} +\textbf{e},
\end{equation}
where $\textbf{x}$ (ground truth signal) can be a scalar/vector drawn from distribution $\mathcal{P}_\textbf{x}$, $\textbf{e}$ is addictive Gaussian noise, $\lambda\geq 0$ is the signal-to-noise ratio, and $\textbf{y}$ is the observed variable. Our goal is to study how $\Delta(\textbf{y} \rightarrow \textbf{x})/\Delta(\textbf{x} \rightarrow \textbf{y})$ and its bound behave as a function of $\lambda$. 
Following the procedure above, we need to find the joint and marginal distributions of random variables $r_\textbf{x}$ and $r_\textbf{y}$ denoting the pairwise distances in ground $\textbf{x}$ and $\textbf{x}$, which are defined as 
\begin{equation*}
    r_\textbf{x} = ||\textbf{x}-\textbf{x}'|| \quad\text{and}\quad   r_\textbf{y} = ||\textbf{y}-\textbf{y}'||.
\end{equation*}
Without loss of generality, let's $\textbf{x}, \textbf{x}'\sim \mathcal{N}(0, \mu^2\mathbb{I})$ and $\textbf{e}, \textbf{e}'\sim \mathcal{N}(0, \sigma^2\mathbb{I})$
Define
\begin{equation*}
    \textbf{u} = \textbf{x} - \textbf{x}' \quad \text{and} \quad     \textbf{v} = \textbf{e} - \textbf{e}',
\end{equation*}then $   r_\textbf{x} = ||\textbf{u}||$ and $r_\textbf{y} = ||\textbf{u}+\textbf{v}||$. Our goal is to compute
\begin{align*}
    p(r_\textbf{x}) &= \mathbb{E}_{\textbf{u}}[\delta(||\textbf{u}||- r_\textbf{x})],\\ 
     p(r_\textbf{y}) &= \mathbb{E}_{\textbf{u},\textbf{v}}[\delta(||\textbf{u}+\textbf{v}||- r_\textbf{y})]\\
      p(r_\textbf{x}, r_\textbf{y}) &= \mathbb{E}_{\textbf{u},\textbf{v}}[\delta(||\textbf{u}||- r_\textbf{x})\delta(||\textbf{u}||- r_\textbf{x})]
\end{align*}.
Since $\textbf{u}\sim\mathcal{N}(0, 2\mu^2\mathbb{I})$, its probability density function is:
\begin{equation*}
    f(u) = (4 \pi \mu^2)^{-d/2} \exp{\left( -\frac{||\textbf{u}||^2}{4\mu^2}\right)}.
\end{equation*}
In spherical coordinates, can be written as $\textbf{u} = r\omega$, with $r\geq 0$, $\omega\in S^{d-1}$. Therefore
\begin{align*}
    p(r_\textbf{x}) &= \int(\textbf{u})\delta(||\textbf{u}||- r_\textbf{x})d\textbf{u}= \int_{0}^{\infty}\int_{S^{d-1}}f(r\omega)\delta(r- r_\textbf{x})r^{d-1}drd\Omega \\&= \int_{S^{d-1}}f(r_\textbf{x}\omega)r_\textbf{x}^{d-1}d\Omega = \int_{S^{d-1}}d\Omega f(r_\textbf{x}\omega)r^{d-1},
\end{align*}
where $\int_{S^{d-1}}d\Omega$ is the surface area of the unit sphere in $\mathbb{R}^d$ that is equal to 
\begin{equation*}
    A_{d-1}= \int_{S^{d-1}}d\Omega = \frac{2\pi^{d/2}}{\Gamma(\frac{d}{2})}.
\end{equation*} Thus,
\begin{equation}
    p(r_\textbf{x}) = \frac{1}{2^{d-1}\mu^d \Gamma(\frac{d}{2})}r_\textbf{x}^{d-1}\exp{\left(-\frac{r_\textbf{x}^2}{4 \mu^2}\right)}.
\end{equation}
Similarly, 
\begin{equation}
    p(r_\textbf{y}) = \frac{1}{2^{d-1}(\mu^2 + \sigma^2)^{d/2} \Gamma(\frac{d}{2})}r_\textbf{y}^{d-1}\exp{\left(-\frac{r_\textbf{y}^2}{4 (\mu^2 + \sigma^2)}\right)}.
\end{equation}
The joint probability distribution
\begin{align*}
    p(r_\textbf{x}, r_\textbf{y}) &= \mathbb{E}_{\textbf{u},\textbf{v}}[\delta(||\textbf{u}||- r_\textbf{x})\delta(||\textbf{u}||- r_\textbf{x})]\\
     &=\int \int f(\textbf{u})f(\textbf{v})\delta(||\textbf{u}||- r_\textbf{x})\delta(||\textbf{u} + \textbf{v}||- r_\textbf{y})d\textbf{u} d\textbf{v}
     \\
     &=\int f(\textbf{u})\delta(||\textbf{u}||- r_\textbf{x}) \left[ f(\textbf{w}- \textbf{u})\int \delta(||\textbf{w}||- r_\textbf{y})d\textbf{w}\right] d\textbf{u}
     \\
     &=\int f(\textbf{u})\delta(||\textbf{u}||- r_\textbf{x}) \left[ \int_{S^{d-1}} f(r_\textbf{y}\omega_\textbf{y} - \textbf{u})r_\textbf{y}^{d-1} d\Omega(\omega_\textbf{y})\right] d\textbf{u}
     \\
     &=\int_{S^{d-1}}\int_{S^{d-1}}f(r_\textbf{x}\omega_\textbf{x}) f(r_\textbf{y}\omega_\textbf{y} - r_\textbf{x}\omega_\textbf{x})r_\textbf{x}^{d-1} r_\textbf{y}^{d-1} d\Omega(\omega_\textbf{x})d\Omega(\omega_\textbf{y})
     \\
     &=\frac{(r_\textbf{x}r_\textbf{y})^{d-1}}{(4\pi \mu^2)^d} \exp{\left(-\frac{r_\textbf{x}^2}{4\mu^2}\right)}\int_{S^{d-1}}\int_{S^{d-1}} \exp{\left(- \frac{||r_\textbf{y} \omega_\textbf{y}- r_\textbf{x}\omega_\textbf{x}||^2}{4\sigma^2} \right)} d \Omega(\omega_\textbf{x}) d\Omega(\omega_\textbf{y})
     \\
     &=\frac{(r_\textbf{x}r_\textbf{y})^{d-1}}{(4\pi \mu^2)^d} \exp{\left(-\frac{r_\textbf{x}^2}{4\mu^2} - \frac{(\mu^2 + \sigma^2)r_\textbf{x}^2}{4 \mu^2 \sigma^2}\right)}\int_{S^{d-1}}\int_{S^{d-1}} \exp{\left(- \frac{r_\textbf{y}r_\textbf{x} \omega_\textbf{y}^T\omega_\textbf{x}}{2\sigma^2} \right)} d \Omega(\omega_\textbf{x}) d\Omega(\omega_\textbf{y})
\end{align*}
Let $c = r_\textbf{x}r_\textbf{y}/2\sigma^2$ and $I$ represent the double spherical integral.
\begin{equation*}
    I = \int_{S^{d-1}}\int_{S^{d-1}} \exp{\left(- c \omega_\textbf{y}^T\omega_\textbf{x} \right)} d \Omega(\omega_\textbf{x}) d\Omega(\omega_\textbf{y})
\end{equation*}
Due to the symmetric properties of a sphere, let $w_\textbf{y} = \textbf{e}$ be fixed, then  
\begin{equation*}
    I = A_{d-1}\int_{S^{d-1}} \exp{\left(- c \textbf{e}^T\omega_\textbf{x} \right)} d \Omega(\omega_\textbf{x})
\end{equation*}, where $A_{d-1}= 2\pi ^{d/2}/\Gamma(d/2)$. The integral $\int_{S^{d-1}} \exp{\left(- c \textbf{e}^T\omega_\textbf{x} \right)} d \Omega(\omega_\textbf{x})$ is the normalization constant of the von Mises-Fisher (vMF) distribution \cite{banerjee2005clustering} and is given by
\begin{equation}
    \int_{S^{d-1}} \exp{\left(- c \textbf{e}^T\omega_\textbf{x} \right)} d \Omega(\omega_\textbf{x}) = (2\pi)^d/2 c^{1-d/2} \mathcal{I}_{d/2-1}(c),
\end{equation} where $\mathcal{I}(\cdot)$ represents the modified Bessel function of the first kind and order $r$. Therefore, putting everything together, we get
\begin{equation}
     p(r_\textbf{x}, r_\textbf{y}) = \frac{ 1}{2^d \mu^d\sigma^2  \Gamma(d/2)}(r_\textbf{x}r_\textbf{y})^{d-1} \exp{\left(-\frac{r_\textbf{x}^2}{4\mu^2} - \frac{(\mu^2 + \sigma^2)r_\textbf{x}^2}{4 \mu^2 \sigma^2}\right)}\mathcal{I}_{d/2 -1}(r_\textbf{x}r_\textbf{y}/2\sigma^2) 
\end{equation}    
\subsubsection{Restricted Mutual Information for 1-d Embedding Models and Gaussian Denoising Models}
We want to compute the following quantity for the two models we described above. 
\begin{equation*}
 \lim_{\epsilon \rightarrow 0} \frac{\text{I}^\epsilon(r_x \rightarrow r_y)}{\epsilon} = \lim_{\eta  \rightarrow 0} \frac{1}{p(r_x=\eta ) }\int_0^\infty dr_y p(r_x=\eta \;,\;r_y)\log\left(\frac{p(r_x= \eta  \;,\; r_y)}{p(r_x= \eta )\; p(r_y)}\right)
\end{equation*}
Let's begin with the 1-d embedding case, the logarithm of the ratio in the integrand is as follows:
\begin{align*}\log\left( \frac{p(r_x, r_y)}{p(r_x)p(r_y)}\right) &=\log\left(p(r_x, r_y)\right) - \log\left(p(r_x)\right) - \log\left(p(r_y)\right)\\& = -\frac{1}{2}\log(1- \rho^2) - \frac{\rho^2 r_x^2}{4 \sigma^2_x(1 - \rho^2)} - \frac{\rho^2 r_y^2}{4 \sigma^2_y(1 - \rho^2)} + \log\left( \cosh\left( \frac{\rho r_x r_y}{2 \sigma_x \sigma_y (1 - \rho^2)}\right)\right)
\end{align*}
Let us denote the integrand by
\begin{equation*}
    G(r_x, r_y) = p(r_x, r_y)\log\left( \frac{p(r_x, r_y)}{p(r_x)P(r_y)}\right).
\end{equation*}
For small $\eta$, we expand it around $r_x = 0$, and since $ G(r_x, r_y)$ is an even function in $r_x$, the leading term is $ G(r_x = 0, r_y)$.
Thus,
\begin{align*}
    G(r_x =0 , r_y) &= p(r_x = 0, r_y)\log\left( \frac{p(r_x = 0, r_y)}{p(r_x = 0)p(r_y)}\right)\\
    & = \frac{1}{
 \pi \sigma_x \sigma_y \sqrt{(1-\rho^2)}}\exp\left(-\frac{r_y^2}{4\sigma_y^2(1-\rho^2)}\right)\left[ -\frac{1}{2}\log(1- \rho^2)  - \frac{\rho^2 r_y^2}{4 \sigma^2_y(1 - \rho^2)}\right]
\end{align*}
Now, we integrate over $r_y$ from $0$ to $\infty$
\begin{align*}
    \int_0^\infty dr_y G(r_x =0 , r_y)  &=\int_0^\infty dr_y  \frac{\exp\left(-\frac{r_y^2}{4\sigma_y^2(1-\rho^2)}\right)}{
 \pi \sigma_x \sigma_y \sqrt{(1-\rho^2)}}\left[ -\frac{1}{2}\log(1- \rho^2)  - \frac{\rho^2 r_y^2}{4 \sigma^2_y(1 - \rho^2)}\right]\\
 &=  -\frac{\log(1- \rho^2)\sqrt{4 \pi\sigma_y^2 (1 - \rho^2)}}{4\pi \sigma_x \sigma_y \sqrt{(1-\rho^2)}} - \frac{\rho^2\sqrt{4\pi} \sigma_y^3 (1 - \rho^2)^{\frac{3}{2}}}{4 \pi\sigma_x\sigma^3_y(1 - \rho^2)^{\frac{3}{2}} }\\
 &=  \frac{1}{2\sqrt{\pi \sigma_x^2}}(-\log(1- \rho^2) - \rho^2)
 \end{align*}
 Therefore,
 \begin{equation}
    \lim_{\epsilon \rightarrow 0} \frac{\text{I}^\epsilon(r_x \rightarrow r_y)}{\epsilon} = \frac{1}{p(r_x=0) } \int_0^\infty dr_y G(r_x =0 , r_y) = -\frac{1}{2}\log(1 - \rho^2)  - \frac{1}{2}\rho^2
\end{equation}
For the Gaussian denoising models, using the asymptotic approximation of the Bessel function given by
\begin{equation}
    \mathcal{I}_\nu(c) = \frac{1}{\Gamma(\nu +1)}\left( \frac{c}{2}\right)^\nu
\end{equation}
for $c\ll 1$. Following the same recipes for the computation in the case of 1-d embedding, we obtained the following. 
 \begin{equation}
    \lim_{\epsilon \rightarrow 0} \frac{\text{I}^\epsilon(r_x \rightarrow r_y)}{\epsilon}  = \frac{d}{2}\left(\log(1 + \lambda)  - \frac{\lambda}{1 + \lambda}\right).
\end{equation}
\end{document}